\documentclass{article}

\PassOptionsToPackage{numbers,sort&compress}{natbib}
 \usepackage[dblblindworkshop, final]{neurips_2026}
\workshoptitle{Physical World AI: Geometry, Characteristics, and Multimodal Sensing}

\usepackage[utf8]{inputenc}
\usepackage[T1]{fontenc}
\usepackage{hyperref}
\usepackage{url}
\usepackage{booktabs}
\usepackage{amsmath,amssymb,amsfonts}
\usepackage{graphicx}
\usepackage{microtype}
\usepackage{xcolor}

\title{Toward a Multimodal Foundation Model\\
for Plasma State Reconstruction}

\author{%
  Bin Xia \\
  School of Physics \\
  Georgia Institute of Technology \\
  Atlanta, GA 30332, USA \\
  \texttt{xiabin@gatech.edu} \\
  \And
  Chuanfei Dong \\
  Department of Astronomy \\
  Department of Electrical and Computer Engineering \\
  Boston University \\
  Boston, MA 02215, USA \\
  \And
  Boting Li \\
  Princeton Plasma Physics Laboratory \\
  Princeton, NJ 08540, USA \\
  \And
  Yilan Qin \\
  Department of Electrical and Computer Engineering \\
  Boston University \\
  Boston, MA 02215, USA \\
  \And
  Zikang Xie \\
  Department of Astronomy \\
  Boston University \\
  Boston, MA 02215, USA \\
  \And
  Sung Hyun Son \\
   Department of Astrophysical Sciences \\
  Princeton University \\
  Princeton, NJ 08544, USA \\
  Princeton Plasma Physics Laboratory \\
  Princeton, NJ 08540, USA \\
  \And
  Adam J. Stanier \\
  Los Alamos National Laboratory \\
  Los Alamos, NM 87545, USA \\
  \And
  Jongsoo Yoo \\
  Princeton Plasma Physics Laboratory \\
  Princeton, NJ 08540, USA \\
  \And
  Hantao Ji \\
  Department of Astrophysical Sciences \\
  Princeton University \\
  Princeton, NJ 08544, USA \\
  Princeton Plasma Physics Laboratory \\
  Princeton, NJ 08540, USA \\
  \And
  Ahmed Diallo \\
  Princeton Plasma Physics Laboratory \\
  Princeton, NJ 08540, USA \\
  \And
  Han Ding \\
  Department of Mechanical Engineering \\
  Carnegie Mellon University \\
  Pittsburgh, PA 15213, USA \\
  \And
  John Wise \\
  School of Physics \\
  Georgia Institute of Technology \\
  Atlanta, GA 30332, USA
}

\begin{document}

\maketitle

\begin{abstract}
Plasma experiments observe coupled physical states through diagnostics that are sparse, heterogeneous, and incomplete in both space and time. We present a proof-of-concept toward a multimodal foundation model for plasma state reconstruction during magnetic reconnection, motivated by the diagnostic environment of the Facility for Laboratory Reconnection Experiments (FLARE). The model is built from a corpus of 250 VPIC magnetic reconnection simulations at fixed reference electron beta, $\beta_{e0}=0.2$, with 225 complete runs for training and 25 disjoint validation runs. A single 5.5M-parameter residual 3D U-Net receives masked three-component magnetic-field and electron-density inputs together with explicit observation masks, preserves temporal resolution through spatial-only pooling, and uses global spatiotemporal self-attention with rotary positional encoding. Training independently masks the magnetic field and density modalities with spatially random, grid, block, and temporal missingness over a wide range of observation fractions. The resulting model supports multiple reconstruction settings without task-specific retraining. When density is completely hidden, as little as 0.1\% magnetic spatial visibility reduces median density NRMSE from approximately 0.9 to 0.1; conversely, fully observed plasma density substantially improves current density reconstruction when magnetic observations are absent or extremely sparse. With no magnetic observations, roughly 1\% density visibility is sufficient to approach the density reconstruction error obtained with fully observed magnetic fields, although accurate current structure derived from magnetic fields remains much more dependent on direct magnetic measurements. Full magnetic conditioning also stabilizes conditional completion of missing future density frames. These results demonstrate bidirectional cross-modal information transfer and support masked multimodal reconstruction as a promising route toward reusable plasma state estimation models, including future FLARE applications. The present study is limited to synthetic VPIC data at one reference electron beta and does not yet establish sim-to-real or cross-regime generalization.
\end{abstract}

\section{Introduction}

Magnetic reconnection changes magnetic topology and converts stored magnetic energy into plasma flows, heating, and particle energization across laboratory, space, solar, and astrophysical systems \citep{yamada2010reconnection,ji2022reconnection}. The Facility for Laboratory Reconnection Experiments (FLARE) at the Princeton Plasma Physics Laboratory is designed to access large-system, multiple-X-line reconnection regimes that are difficult to realize in smaller laboratory devices \citep{ji2026flare}. Its experimental program is therefore naturally associated with a state-estimation problem: the underlying plasma evolves as a coupled spatiotemporal system, while diagnostics provide only partial and modality-dependent observations.

A useful reconstruction model for such experiments should not require a separate network for every diagnostic configuration. It should accept diverse subsets of available measurements, exploit correlations between physical quantities, and reconstruct the unobserved state over both space and time. Existing machine learning approaches have demonstrated strong performance for sparse field reconstruction in fluids and geophysical systems \citep{fukami2021voronoi,santos2023senseiver}, and neural networks have accelerated equilibrium reconstruction in fusion plasmas from magnetic and non-magnetic measurements \citep{lu2023east,bonotto2024planet}. However, these systems are usually developed around a fixed sensor representation or a task-specific mapping. In parallel, masked modeling has emerged as a general self-supervised objective for learning reusable image, video, and multimodal representations \citep{he2022mae,tong2022videomae,bachmann2022multimae}, and scientific foundation models have begun to exploit heterogeneous physical variables for multiple downstream tasks \citep{nguyen2023climax}.

Here we combine these ideas into a single masked spatiotemporal reconstruction model for plasma fields. We treat the three-component magnetic field $\mathbf{B}=(B_x,B_y,B_z)$ and normalized electron density $\rho$ as two coupled modalities. The processed notation $(B_x,B_y,B_z)$ corresponds to the native cylindrical VPIC components $(B_R,B_\phi,B_Z)$, and $\rho$ denotes $n_e/n_0$ before standardization. During training, each modality is independently subjected to spatially random, regularly sparse, contiguous-block, or temporal missingness with observation fractions ranging from fully absent to fully observed. A single residual 3D U-Net is trained to reconstruct the complete four-field state from the masked fields and their binary observation masks.

Our study is a prototype rather than a claim that a broadly general plasma foundation model has already been achieved. The present model is small by modern foundation-model standards, is trained only on synthetic VPIC data at $\beta_{e0}=0.2$, and has not yet been adapted to FLARE measurements. We therefore use ``foundation model'' only in a limited operational sense: one set of weights is trained with a broad masked objective and reused across distinct observation geometries, modality combinations, super-resolution settings, and temporal reconstruction tasks. This framing does not imply cross-regime, cross-device, or simulation-to-experiment generalization, none of which is established here. The main contributions are:
\begin{itemize}
    \item a unified mask-conditioned spatiotemporal reconstruction objective for magnetic field and density modalities with independently sampled missingness;
    \item a residual 3D U-Net with preserved temporal resolution and global spatiotemporal attention that can infer the full state without hard-clamping observed values;
    \item controlled experiments demonstrating bidirectional transfer between magnetic and density modalities, including magnetic ablation, density super-resolution, and conditional temporal completion; and
    \item a sliding-window procedure showing how the same 24-frame model can be applied to longer plasma sequences.
\end{itemize}

\section{Related work and background}

\paragraph{Magnetic reconnection and FLARE.}
Magnetic reconnection is a multiscale process whose global evolution can depend on kinetic dissipation, current-sheet instability, plasmoid formation, and coupling between microscopic and macroscopic scales \citep{yamada2010reconnection,ji2022reconnection}. FLARE was constructed specifically to study reconnection in multiple-X-line regimes relevant to space, astrophysical, and fusion plasmas \citep{ji2026flare}. This setting motivates data-driven methods that can fuse heterogeneous diagnostics while respecting the strongly coupled spatial and temporal structure of the plasma.

\paragraph{Machine learning for sparse physical reconstruction.}
Data-driven reconstruction from sparse sensors has a long history in fluid mechanics and scientific machine learning \citep{brunton2020mlfluid,karniadakis2021piml}. Voronoi-assisted convolutional reconstruction can accommodate arbitrary sensor locations \citep{fukami2021voronoi}, while the attention-based Senseiver handles variable-size sparse observation sets and complex fields \citep{santos2023senseiver}. In plasma physics, deep neural networks have been used to infer tokamak equilibrium quantities from diagnostic measurements \citep{lu2023east}, and PlaNet combines magnetic and non-magnetic constraints with a physics-informed neural operator for equilibrium and separatrix reconstruction \citep{bonotto2024planet}. Our focus differs in both geometry and objective: rather than learn one fixed mapping from diagnostics to an equilibrium descriptor, we train one network to recover a time-dependent multi-field state under a distribution of missing-data patterns.

\paragraph{Masked multimodal representation learning.}
Masked autoencoding learns representations by reconstructing deliberately removed content \citep{he2022mae}; VideoMAE extends the principle to space-time inputs \citep{tong2022videomae}. MultiMAE showed that masking across multiple input and output modalities can induce cross-modal predictive coding and allow the same model to operate under different modality subsets \citep{bachmann2022multimae}. ClimaX demonstrated a related foundation-model perspective for heterogeneous physical variables in weather and climate \citep{nguyen2023climax}. Our approach follows this general principle but operates directly on plasma field tensors and explicitly samples physically motivated spatial and temporal observation patterns.

\section{Data and problem formulation}

\subsection{VPIC reconnection corpus}

We use a synthetic corpus generated with the VPIC particle-in-cell code \citep{bowers2008vpic}. The present work uses the $\beta_{e0}=0.2$ slice of a two-dimensional axisymmetric reconnection scan, comprising 250 simulations spanning
$\nu\in\{0,1,2,3,5,6,7,8,9,10\}$,
$B_{z0}/B_0\in\{0,0.1,0.15,0.2,0.3\}$, and
$\tau\omega_{ci0}\in\{40,70,100,150,200\}$.
Here $\beta_{e0}$ is the reference electron beta; full simulation geometry, normalization, drive, collision model, and parameter definitions are given in Appendix~\ref{app:vpic}.

The native VPIC outputs $(B_R,B_\phi,B_Z,n_e/n_0)$ are represented in the machine-learning arrays as $(B_x,B_y,B_z,\rho)$, with $\rho=n_e/n_0$. The retained domain is $30\le R\le130$~cm and $|Z|\le21$~cm; using $x\equiv R-80~\mathrm{cm}$ gives a processed grid of $(X,Z)=(154,62)$ spanning $x\in[-50,50]$~cm and $z\in[-21,21]$~cm. We extract contiguous 24-frame windows with a start-time stride of two frames. The split is performed by complete simulation run rather than by window: 225 runs provide 7,059 training windows and 25 disjoint validation runs provide 591 validation windows, so overlapping windows from the same physical trajectory never appear in both sets.

Each channel is standardized using mean and standard deviation statistics estimated from the training data only. The resulting statistics are
\[
(\mu,\sigma)_{B_x}=(0.0029,0.1962),\quad
(\mu,\sigma)_{B_y}=(0.0099,0.0656),
\]
\[
(\mu,\sigma)_{B_z}=(-0.0123,0.2395),\quad
(\mu,\sigma)_{\rho}=(0.5605,0.3284).
\]
The current corpus is a methodological proxy for FLARE-oriented reconstruction rather than a validated digital twin of FLARE. Cross-$\beta_{e0}$ generalization and adaptation to experimental measurements are left for future work.

\subsection{Sparse and incomplete observations}

Let $\mathbf{M}_{B}\in\{0,1\}^{1\times T\times X\times Z}$ denote the common observation mask for the three magnetic components and $\mathbf{M}_{\rho}$ the density mask. The observed standardized fields are
\[
    \tilde{\mathbf{x}}_{\rm obs}
    =
    [\,\mathbf{M}_{B}\odot \tilde{\mathbf{B}},
       \mathbf{M}_{\rho}\odot\tilde{\rho}\,].
\]
The network input concatenates the four masked fields with four binary mask channels, using the same $\mathbf{M}_{B}$ for $B_x,B_y,B_z$. Explicit masks distinguish a physically zero value from a missing value.

During training, magnetic and density modalities independently sample one of five equally weighted families: spatial random, spatial grid, spatial block, temporal random, and temporal block. Random/grid spatial masks use a three-part visibility mixture: 5\% probability of no visible samples, 5\% probability of complete visibility, and otherwise an exact visible count sampled approximately log-uniformly between 1 and $XZ-1$. Spatial blocks sample a random area and geometry, with equal probability of inpainting (hidden interior) and outpainting (visible interior). Temporal patterns sample the number of visible frames from $0$ to $T$; temporal random uses scattered full frames, whereas temporal block uses a contiguous hidden or visible interval. Magnetic and density pattern, severity, and layout are independently drawn.

\section{Method}

\subsection{Residual spatiotemporal U-Net}

The model receives an 8-channel tensor,
\[
  [\tilde{B}_{x,\rm obs},\tilde{B}_{y,\rm obs},\tilde{B}_{z,\rm obs},
  \tilde{\rho}_{\rm obs},
  M_{B},M_{B},M_{B},M_{\rho}],
\]
and predicts all four complete standardized fields. The backbone is a four-level residual 3D U-Net \citep{ronneberger2015unet} with channel widths $(24,48,96,192)$ and 5,526,100 trainable parameters. Each residual convolution block applies a $3\times3\times3$ convolution, SiLU, and a second $3\times3\times3$ convolution, followed by addition of a strictly linear skip path; a $1\times1\times1$ projection is used only when the channel count changes. We intentionally omit activation normalization in the convolutional path so that absolute feature levels and amplitudes can propagate through the linear residual path.

All downsampling is spatial only, using max pooling with kernel and stride $(1,2,2)$, so the temporal dimension remains 24 throughout the encoder. Decoder features are trilinearly upsampled to the corresponding encoder resolution, concatenated with U-Net skip features, and reduced through the same residual convolution blocks. Full stage-by-stage shapes and parameter counts are given in Appendix~\ref{app:architecture}.

\subsection{Global spatiotemporal attention}

We insert global self-attention \citep{vaswani2017attention} after the third encoder feature map (96 channels at $24\times38\times15$) and after the bottleneck residual block (192 channels at $24\times19\times7$). Flattening the joint $(T,X,Z)$ lattice produces 13,680 and 3,192 tokens, respectively, so each attention layer can route information jointly across space and time rather than along a preselected axis.

Both attention layers use eight heads. The 96-channel layer therefore has head dimension 12 and a $96\!\rightarrow\!288$ joint QKV projection, while the 192-channel bottleneck has head dimension 24 and a $192\!\rightarrow\!576$ QKV projection. The attention module is pre-normalized: LayerNorm is applied before QKV projection, followed by non-causal scaled dot-product self-attention, a linear output projection, and residual addition to the original tokens. There is no separate MLP/FFN sublayer in the attention module. The implementation calls PyTorch scaled-dot-product attention directly rather than constructing an explicit attention matrix or hard-coding a FlashAttention backend; attention dropout is zero.

Rotary position embedding \citep{su2024roformer} is applied to $Q$ and $K$ separately along the temporal, $x$, and $z$ coordinates. At the 96-channel layer, each 12-dimensional head allocates four rotary dimensions to each axis; at the 192-channel bottleneck, each 24-dimensional head allocates eight dimensions to each axis. Thus all head dimensions participate in axial 3D RoPE in the present configuration. Layer normalization remains confined to the attention branch, and the attention output projection is initialized to zero so that each attention module initially contributes an exact zero correction to the convolutional backbone.

\subsection{Residual field prediction and objective}

The final $1\times1\times1$ convolution predicts a correction to the visible standardized fields:
\begin{equation}
    \hat{\mathbf{x}}
    =
    \tilde{\mathbf{x}}_{\rm obs}
    +
    f_\theta(\tilde{\mathbf{x}}_{\rm obs},\mathbf{M}).
\end{equation}
The output convolution is initialized to zero. Therefore, before learning begins, the network returns the supplied standardized values at observed locations and zero at hidden locations, where zero corresponds to the training-set channel mean after standardization. The learned correction is applied everywhere. We intentionally do not use a mask-gated hard data-consistency step: observed sites remain part of the reconstruction objective and can be revised by the model rather than being overwritten after inference.

The loss balances magnetic and density modalities and prevents either the large hidden region or the three magnetic channels from dominating the objective. For each modality group $m\in\{B,\rho\}$ we compute MSE separately over visible and hidden elements and average only the non-empty terms:
\begin{equation}
    \mathcal{L}
    =
    \frac{1}{2}\left(\mathcal{L}_{B}+\mathcal{L}_{\rho}\right),\qquad
    \mathcal{L}_{m}
    =
    \frac{1}{|\mathcal{S}_{m}|}
    \sum_{s\in\mathcal{S}_{m}}\mathrm{MSE}_{m,s},
\end{equation}
where $\mathcal{S}_{m}\subseteq\{\mathrm{visible},\mathrm{hidden}\}$ contains only subsets with at least one element. For the magnetic group, squared errors from $B_x$, $B_y$, and $B_z$ are pooled together before the visible/hidden averages are formed; density is treated as the second group. The final arithmetic mean of $\mathcal{L}_{B}$ and $\mathcal{L}_{\rho}$ therefore gives the magnetic field group and density equal overall weight regardless of channel count or visible fraction.

The final model is trained from random initialization for 4,500 epochs with AdamW, learning rate $2\times10^{-4}$, weight decay $10^{-4}$, a 10-epoch linear warmup, and cosine decay to $2\times10^{-6}$. Training uses automatic mixed precision and distributed data parallelism on 16 NVIDIA A100-SXM4-80GB GPUs on NERSC Perlmutter with per-GPU batch size 8 and global batch size 128. A full validation pass is performed after every training epoch, evaluating all five standardized mask patterns on the held-out validation loader. Training over 4,500 epochs required approximately 12.5 hours of wall-clock time, or about 200 GPU-hours.

\section{Experiments and results}

Unless otherwise stated, all reported results use the final checkpoint at epoch 4,500 (\texttt{latest.pt}), with no task-specific fine-tuning or checkpoint selection. For each setting, window-level metrics are averaged within each validation run; we then report the median and 16th--84th percentiles across the 25 runs. Density NRMSE denotes the RMSE in training-standardized density units, equivalently the density RMSE normalized by the training-set standard deviation.

For magnetic structure we report the toroidal (out-of-plane) current density. In the native cylindrical geometry this is $J_\phi$; to remain consistent with the processed Cartesian-like array notation and existing figures, we denote the same quantity by $J_y$:
\begin{equation}
    J_y\equiv J_\phi =
    \frac{1}{\mu_0}
    \left(
        \frac{\partial B_R}{\partial Z}
        -
        \frac{\partial B_Z}{\partial R}
    \right)
    =
    \frac{1}{\mu_0}
    \left(
        \frac{\partial B_x}{\partial z}
        -
        \frac{\partial B_z}{\partial x}
    \right),
    \qquad
    \mu_0=4\pi\times10^{-7}\ {\rm H\,m^{-1}}.
\end{equation}
After inverse-transforming $B_x$ and $B_z$ to Tesla, we convert the spatial coordinates from centimeters to meters before differentiation. We report $J_y$ in ${\rm A\,m^{-2}}$. The dimensionless $J_y$ NRMSE is the RMSE of the predicted-current residual normalized by the training-set current density standard deviation, $\sigma_{J_y,\rm train}=2.512\times10^6\ {\rm A\,m^{-2}}$. Because the model does not hard-clamp visible magnetic values, $J_y$ error can remain nonzero even when $\mathbf{B}$ is fully observed; spatial differentiation also amplifies small high-frequency field errors.

\subsection{Standardized reconstruction benchmark}

To compare missing-data geometry at matched severity, we evaluate five deterministic approximately 50\%-visible masks shared across all four channels. Table~\ref{tab:standardized} reports run-level median errors with 16th--84th percentile intervals. Overall MSE denotes the unweighted mean squared residual over all four standardized fields and all spatiotemporal locations. Contiguous missing regions produce the largest overall and density errors, whereas scattered and regularly sampled observations are generally easier to reconstruct; $J_y$ error shows additional dependence on sampling geometry.

\begin{table}[t]
    \centering
    \caption{Standardized approximately 50\%-visible evaluation. Entries are median [16th, 84th percentile] across 25 held-out runs.}
    \label{tab:standardized}
    \small
    \begin{tabular}{lccc}
        \toprule
        Task & Overall MSE & Density NRMSE & $J_y$ NRMSE \\
        \midrule
        Spatial random   & 0.0025 [0.0018, 0.0049] & 0.0456 [0.0333, 0.0666] & 0.3173 [0.2535, 0.4796] \\
        Spatial grid     & 0.0033 [0.0023, 0.0062] & 0.0581 [0.0460, 0.0785] & 0.1931 [0.1607, 0.2630] \\
        Spatial block    & 0.0287 [0.0117, 0.0589] & 0.0933 [0.0526, 0.1426] & 0.3724 [0.1888, 0.4926] \\
        Temporal random  & 0.0044 [0.0028, 0.0101] & 0.0405 [0.0288, 0.0613] & 0.1916 [0.1628, 0.2666] \\
        Temporal block   & 0.0217 [0.0066, 0.0576] & 0.0677 [0.0376, 0.1462] & 0.2385 [0.1702, 0.4286] \\
        \bottomrule
    \end{tabular}
\end{table}

\subsection{Spatial reconstruction under 50\% density masking}

\begin{figure}[t]
    \centering
    \includegraphics[width=0.8\linewidth]{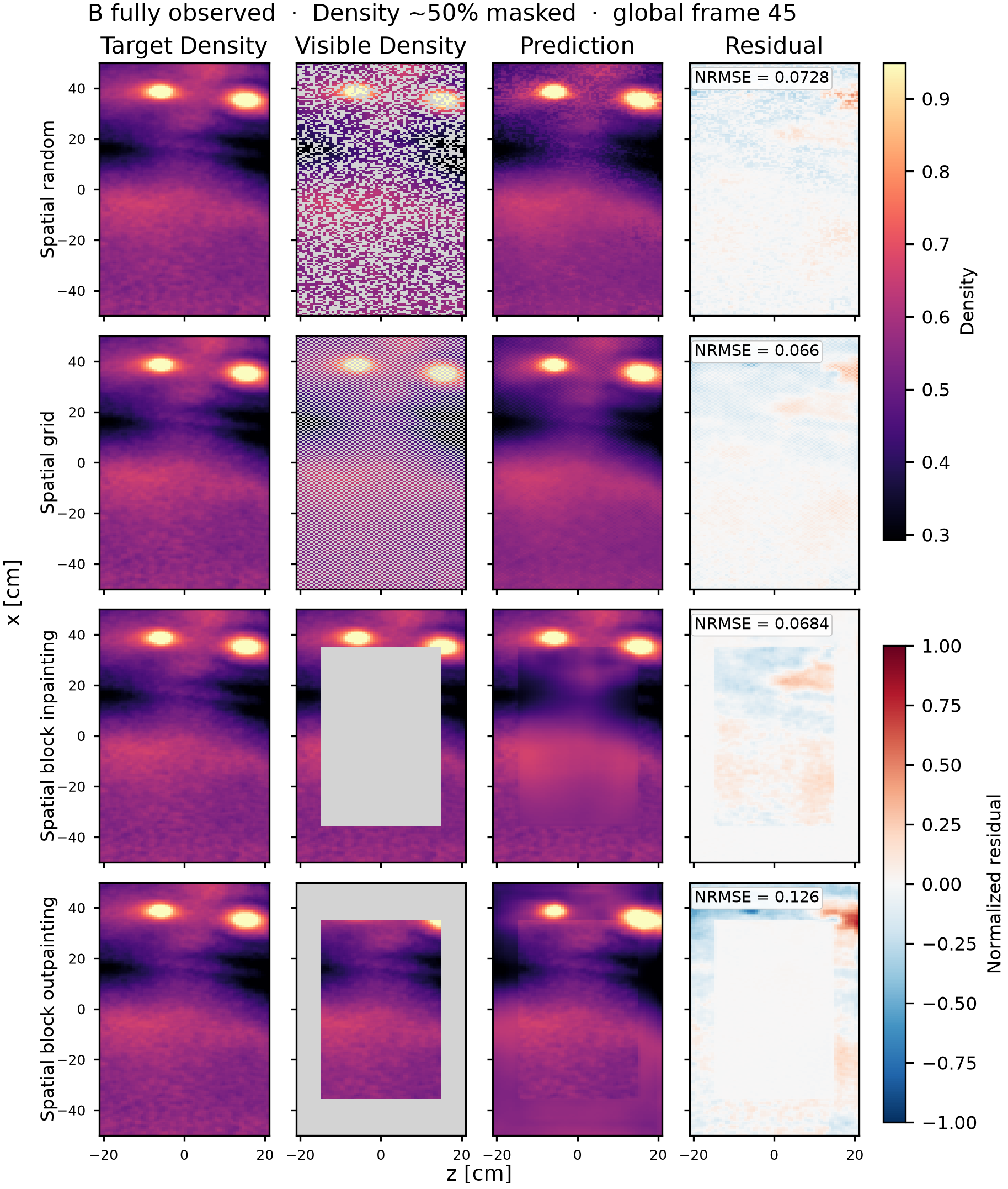}
    \caption{Spatial density reconstruction with the magnetic field fully observed and density approximately 50\% masked. Rows show random sampling, a regular checkerboard, block inpainting, and block outpainting. Prediction panels are raw model outputs rather than hard-clamped fields.}
    \label{fig:spatial}
\end{figure}

Figure~\ref{fig:spatial} compares four spatial density observation geometries for a representative held-out plasmoid-merger frame while the magnetic field is fully observed. Random and checkerboard sampling, as well as block inpainting, recover the dominant density morphology with NRMSE $0.066$--$0.073$. Block outpainting is more difficult (NRMSE $0.126$), because the visible region is confined to the center while the much larger exterior must be inferred. Residuals concentrate around sharp density structures and mask boundaries.

\subsection{Bidirectional cross-modal reconstruction}

\begin{figure}[t]
    \centering
    \includegraphics[width=0.8\linewidth]{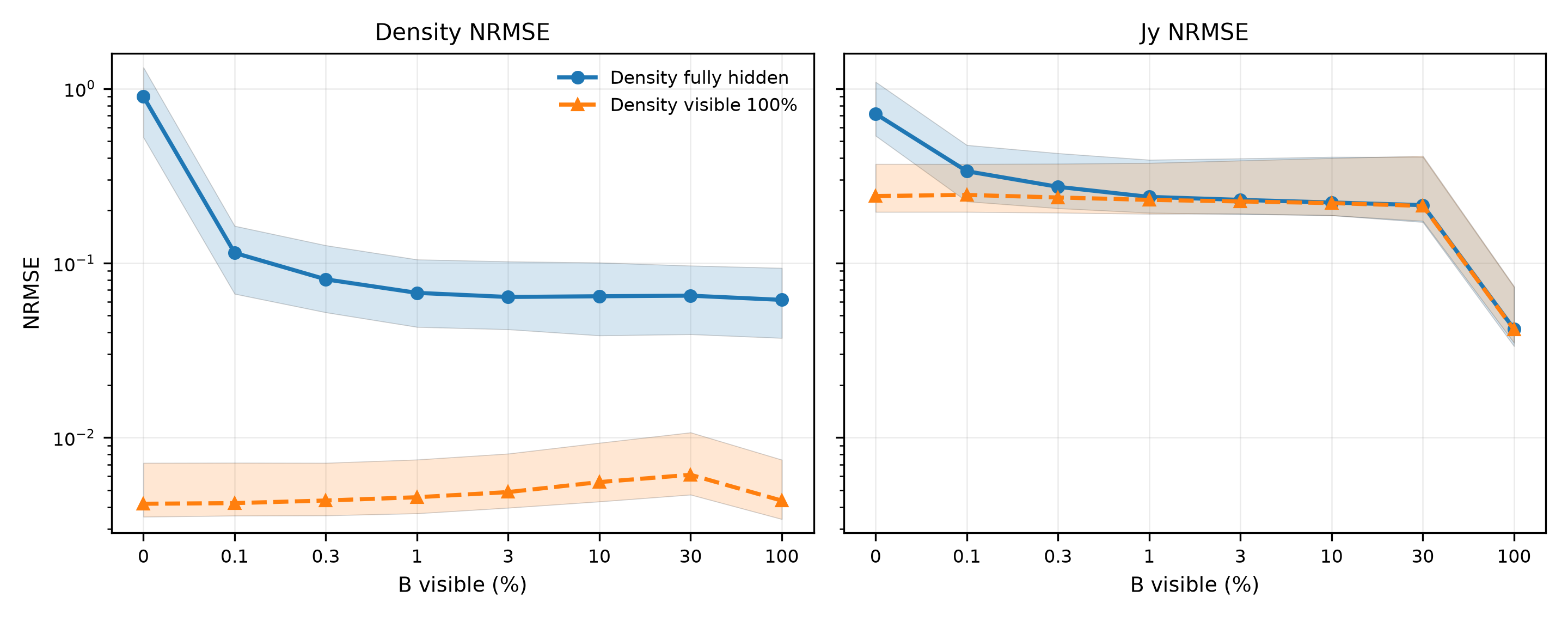}
    \caption{Paired magnetic ablation with nested magnetic probe sets. Blue curves use fully hidden density; orange curves use fully visible density with the identical magnetic masks. Left: density NRMSE. Right: derived $J_y$ NRMSE. Bands denote the 16th--84th percentile across held-out runs.}
    \label{fig:magnetic}
\end{figure}

We next isolate cross-modal information transfer with a paired magnetic ablation (Fig.~\ref{fig:magnetic}). For every held-out window, the two density conditions use exactly the same nested magnetic-probe ranking; only density visibility changes. When density is fully hidden, the median density NRMSE is close to 0.9 with no magnetic observations, but falls to roughly 0.1 with only 0.1\% of spatial magnetic locations visible and approaches a $\sim0.06$--$0.08$ floor around 1\% visibility. Thus sparse magnetic measurements strongly constrain density.

The reverse direction is visible in $J_y$. When magnetic information is absent or extremely sparse, fully observed plasma density substantially reduces current density error relative to the density-hidden case. As magnetic coverage increases, the two $J_y$ curves converge, and at complete magnetic visibility both approach the same low error floor. This paired experiment provides evidence of bidirectional cross-modal predictive benefit rather than independent single-field interpolation.

\subsection{Density super-resolution}

\begin{figure}[t]
    \centering
    \includegraphics[width=0.8\linewidth]{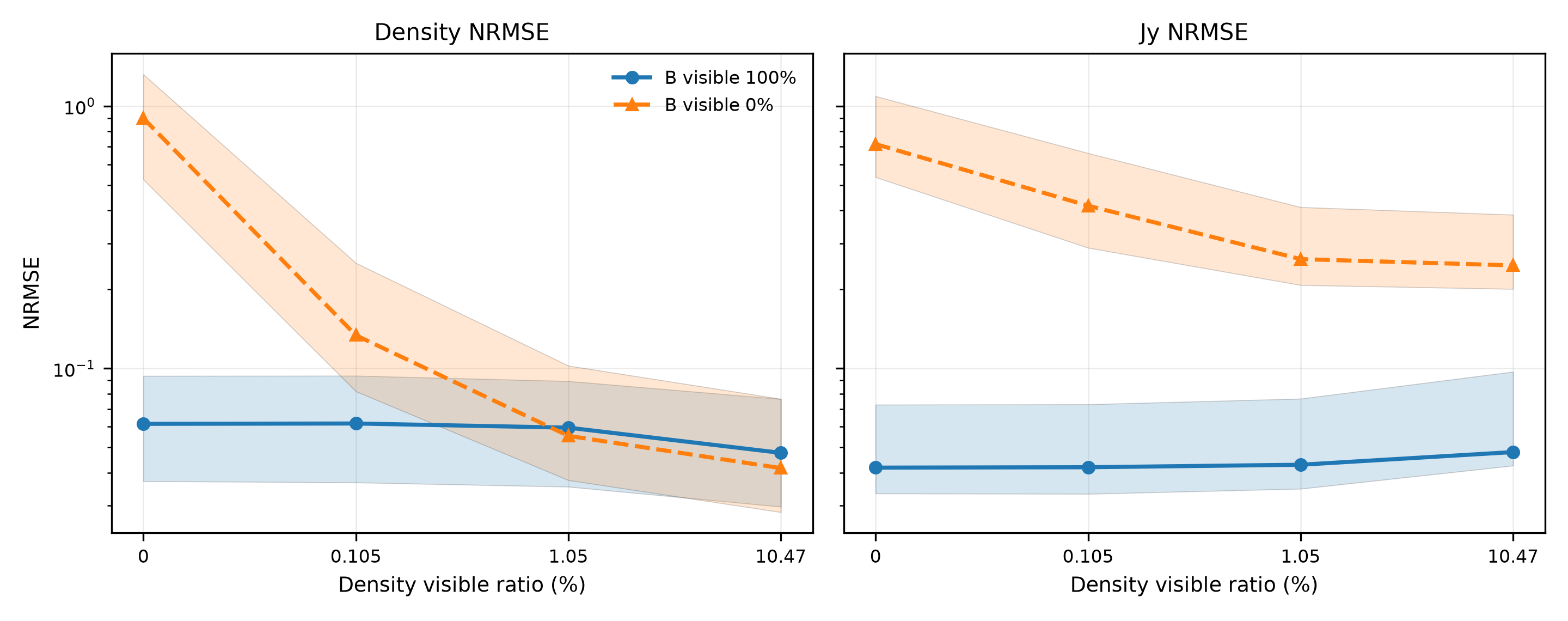}
    \caption{Density super-resolution as a function of exact density probe count, shown as spatial visibility. Fully observed magnetic fields (blue) strongly constrain density even without density probes; when $\mathbf{B}$ is absent (orange), about 1\% density visibility is sufficient to approach the density error floor, while $J_y$ remains more dependent on magnetic observations.}
    \label{fig:superres}
\end{figure}

Figure~\ref{fig:superres} sweeps exact density-probe counts of 0, 10, 100, and 1,000, corresponding to 0, 0.105, 1.05, and 10.47\% of the $154\times62$ spatial grid. With the magnetic field fully observed, density NRMSE is already near 0.06 even with zero density probes, and additional density measurements provide only modest gains. Without magnetic observations, however, density error drops by almost an order of magnitude with only 10 probes and reaches the B-conditioned density error level by roughly 1\% density visibility.

The $J_y$ panel reveals an important asymmetry. Plasma density measurements alone provide information about magnetic and current structure, but even 10.47\% density visibility leaves $J_y$ substantially less accurate than when $\mathbf{B}$ is directly observed. Plasma density is therefore more strongly constrained by cross-modal context than the derived current structure.

\subsection{Conditional temporal completion}

\begin{figure}[t]
    \centering
    \includegraphics[width=0.8\linewidth]{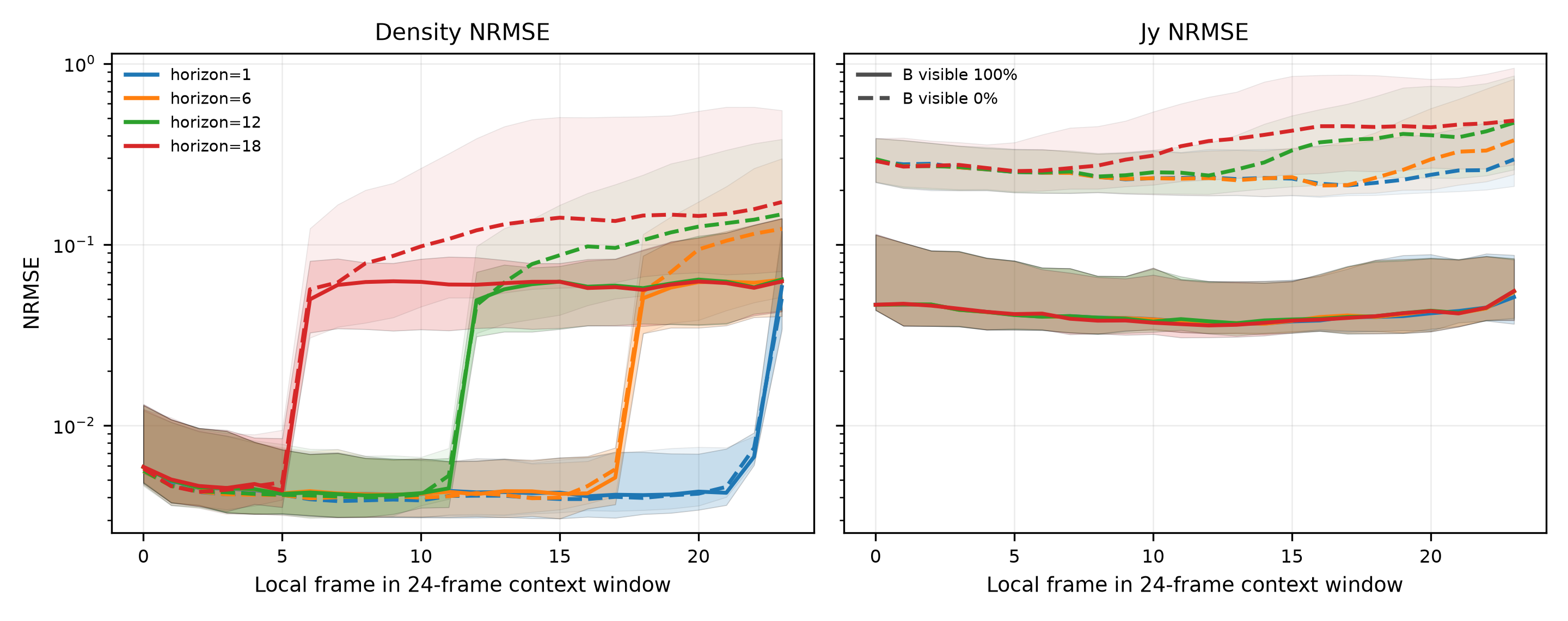}
    \caption{Conditional temporal completion within a 24-frame context. Colors denote the missing-suffix length; solid and dashed curves correspond to fully observed and fully hidden magnetic fields. Full magnetic context stabilizes future density reconstruction, while B-hidden $J_y$ error increases with completion depth.}
    \label{fig:forecast}
\end{figure}

To probe temporal completion, we retain a prefix of the 24-frame density window and hide all later density frames. Prefix lengths of 23, 18, 12, and 6 leave missing suffixes of 1, 6, 12, and 18 frames, respectively. Figure~\ref{fig:forecast} shows that density error remains at a few $10^{-3}$ within the observed prefix and rises sharply at the completion boundary. Longer horizons are more difficult, particularly when magnetic fields are also hidden. With fully observed magnetic fields, future density NRMSE remains near $0.05$--$0.07$ over much of the missing interval; without $\mathbf{B}$, median errors and run-to-run variation grow more strongly with completion depth.

When $\mathbf{B}$ is fully visible, this is conditional temporal completion rather than causal forecasting: the model estimates $\rho_{\rm future}$ conditioned on $\rho_{\rm past}$ and $\mathbf{B}$ over the full 24-frame window. The B-hidden condition is closer to forecasting from density history alone. The substantially larger B-hidden $J_y$ error further shows that missing entire magnetic snapshots is more demanding than interpolating density from cross-modal context.

\section{Discussion}

The experiments support three main conclusions. First, one masked model handles irregular sparse sensing, regular super-resolution, contiguous inpainting/outpainting, modality ablation, and temporal completion without task-specific retraining; this is the limited sense in which the model is foundation-like. Second, magnetic field and density provide complementary information: sparse magnetic observations strongly constrain plasma density, while plasma density helps recover current structure when magnetic measurements are absent or extremely sparse, although $J_y$ remains more dependent on direct magnetic information. Third, missingness geometry matters: scattered spatial and temporal observations are generally easier to reconstruct than contiguous missing blocks. Global attention is intended to route information over long distances, but it does not impose a physical evolution law.

Important limitations remain. Results use only synthetic VPIC simulations at reference electron beta $\beta_{e0}=0.2$, with no demonstrated cross-$\beta_{e0}$ generalization, robustness to experimental noise or calibration error, or sim-to-real transfer to FLARE. The held-out runs are randomly drawn from the same Cartesian parameter grid and form a validation split rather than a separate third test set, so the present evaluation does not test generalization to unseen values of $\nu$, $B_{z0}$, or $\tau$. The masks also represent idealized grid-based observations rather than realistic FLARE diagnostic geometries such as fixed probe layouts or line-integrated measurements. Only magnetic field and density are modeled, and the 5.5M-parameter network is a proof of concept rather than a large-scale foundation model. We also do not isolate the contributions of the residual output formulation, spatial-only pooling, global attention, or RoPE through systematic architecture ablations, and we do not compare against classical sparse-reconstruction baselines such as linear or Voronoi-assisted interpolation, so the reported errors quantify absolute reconstruction quality rather than improvement over simpler methods. Because observed values are not hard-clamped, predictions can deviate from measurements at observed sites, by a few $10^{-3}$ in standardized units in the cases examined here (Fig.~\ref{fig:forecast}). The model does not explicitly enforce physical consistency constraints and produces deterministic point estimates without calibrated predictive uncertainty. Future work should expand the corpus and modalities, test stronger held-out parameter splits and realistic diagnostic geometries, incorporate stronger temporal and physical inductive biases, and evaluate explicit data-consistency and probabilistic objectives.

\section{Conclusion}

We presented a unified masked spatiotemporal model as a step toward a multimodal foundation model for plasma state reconstruction across sparse sensing, cross-modal completion, super-resolution, and temporal completion. The results demonstrate bidirectional predictive coupling between magnetic-field and density observations within a single model. This single-$\beta_{e0}$ synthetic study is a proof of concept; cross-regime and simulation-to-experiment transfer, including FLARE, remain future work.

\begin{ack}
This research used resources of the National Energy Research Scientific Computing Center (NERSC), a U.S. Department of Energy Office of Science User Facility, under NERSC award GenAI@NERSC. C.D. acknowledges support from the U.S. Department of Energy under Grant No. DE-SC0024639 and the Alfred P. Sloan Research Fellowship.
\end{ack}

\newpage
\bibliographystyle{plainnat}
\bibliography{references}

@article{yamada2010reconnection,
  author  = {Yamada, Masaaki and Kulsrud, Russell and Ji, Hantao},
  title   = {Magnetic reconnection},
  journal = {Reviews of Modern Physics},
  volume  = {82},
  number  = {1},
  pages   = {603--664},
  year    = {2010},
  doi     = {10.1103/RevModPhys.82.603}
}

@article{ji2022reconnection,
  author  = {Ji, Hantao and Daughton, William and Jara-Almonte, Jonathan and Le, Ari and Stanier, Adam and Yoo, Jongsoo},
  title   = {Magnetic reconnection in the era of exascale computing and multiscale experiments},
  journal = {Nature Reviews Physics},
  volume  = {4},
  number  = {4},
  pages   = {263--282},
  year    = {2022},
  doi     = {10.1038/s42254-021-00419-x}
}

@misc{ji2026flare,
  title         = {The {FLARE} Facility},
  author        = {Ji, Hantao and Yoo, Jongsoo and Shi, Peiyun and Jung, Euichan and Maheshwari, Kush and Robbins, Adam and Son, Sunghyun and Stanier, Adam and Ren, Yang and Bose, Sayak and Corl, Dylan and Corrigan, Keith and Cutler, Robert and Daughton, William and Ellis, Robert and Gettelfinger, Geoffrey and Hatcher, Ronald and Heitzenroeder, Philip and Hoffmann, Frank and Jara-Almonte, Jonathan and Kalish, Michael and Kozub, Thomas and Merino, Enrique and Que, Weiguo and Smith, Benjamin and Wallace, John and Zhao, Xin and Avrutsky, Sofia and Crocker, Neal and Dorfman, Seth and Doke, Yuka and Dunn, Corina and Fox, William and Li, Boting and Li, Xiaocan and Myers, Clayton and Okazaki, Shohgo and Pawlak, Joshua and Rhee, Tongnyeol and Seo, Byonghoon and Suzuki, Taiju and Zhang, Yang and Bale, Stuart and Bhattacharjee, Amitava and Carter, Troy and Dong, Chuanfei and Drake, James and Egedal, Jan and Gilson, Erik and Inomoto, Michiaki and Menard, Jonathan and Ono, Yasushi and Prager, Stewart and Sarff, John and Tanabe, Hiroshi and Yamada, Masaaki},
  year          = {2026},
  eprint        = {2608.17332},
  archivePrefix = {arXiv},
  primaryClass  = {physics.plasm-ph},
  url           = {https://arxiv.org/abs/2608.17332}
}

@article{bowers2008vpic,
  author  = {Bowers, K. J. and Albright, B. J. and Yin, L. and Bergen, B. and Kwan, T. J. T.},
  title   = {Ultrahigh performance three-dimensional electromagnetic relativistic kinetic plasma simulation},
  journal = {Physics of Plasmas},
  volume  = {15},
  number  = {5},
  pages   = {055703},
  year    = {2008},
  doi     = {10.1063/1.2840133}
}

@article{brunton2020mlfluid,
  author  = {Brunton, Steven L. and Noack, Bernd R. and Koumoutsakos, Petros},
  title   = {Machine Learning for Fluid Mechanics},
  journal = {Annual Review of Fluid Mechanics},
  volume  = {52},
  pages   = {477--508},
  year    = {2020},
  doi     = {10.1146/annurev-fluid-010719-060214}
}

@article{karniadakis2021piml,
  author  = {Karniadakis, George Em and Kevrekidis, Ioannis G. and Lu, Lu and Perdikaris, Paris and Wang, Sifan and Yang, Liu},
  title   = {Physics-informed machine learning},
  journal = {Nature Reviews Physics},
  volume  = {3},
  pages   = {422--440},
  year    = {2021},
  doi     = {10.1038/s42254-021-00314-5}
}

@article{fukami2021voronoi,
  author  = {Fukami, Kai and Maulik, Romit and Ramachandra, Nesar and Fukagata, Koji and Taira, Kunihiko},
  title   = {Global field reconstruction from sparse sensors with Voronoi tessellation-assisted deep learning},
  journal = {Nature Machine Intelligence},
  volume  = {3},
  number  = {11},
  pages   = {945--951},
  year    = {2021},
  doi     = {10.1038/s42256-021-00402-2}
}

@article{santos2023senseiver,
  author  = {Santos, Javier E. and Fox, Zachary R. and Mohan, Arvind and O'Malley, Daniel and Viswanathan, Hari and Lubbers, Nicholas},
  title   = {Development of the Senseiver for efficient field reconstruction from sparse observations},
  journal = {Nature Machine Intelligence},
  volume  = {5},
  number  = {11},
  pages   = {1317--1325},
  year    = {2023},
  doi     = {10.1038/s42256-023-00746-x}
}

@article{lu2023east,
  author  = {Lu, Jingjing and Hu, Youjun and Xiang, Nong and Sun, Youwen},
  title   = {Fast equilibrium reconstruction by deep learning on EAST tokamak},
  journal = {AIP Advances},
  volume  = {13},
  number  = {7},
  pages   = {075007},
  year    = {2023},
  doi     = {10.1063/5.0152318}
}

@article{bonotto2024planet,
  author  = {Bonotto, Matteo and Abate, Domenico and Pigatto, Leonardo},
  title   = {Reconstruction of plasma equilibrium and separatrix using convolutional physics-informed neural operator},
  journal = {Fusion Engineering and Design},
  volume  = {200},
  pages   = {114193},
  year    = {2024},
  doi     = {10.1016/j.fusengdes.2024.114193}
}

@inproceedings{he2022mae,
  author    = {He, Kaiming and Chen, Xinlei and Xie, Saining and Li, Yanghao and Doll{\'a}r, Piotr and Girshick, Ross},
  title     = {Masked Autoencoders Are Scalable Vision Learners},
  booktitle = {Proceedings of the IEEE/CVF Conference on Computer Vision and Pattern Recognition (CVPR)},
  pages     = {16000--16009},
  month     = {June},
  year      = {2022},
  doi       = {10.1109/CVPR52688.2022.01553}
}

@inproceedings{tong2022videomae,
  author    = {Tong, Zhan and Song, Yibing and Wang, Jue and Wang, Limin},
  title     = {{VideoMAE}: Masked Autoencoders Are Data-Efficient Learners for Self-Supervised Video Pre-Training},
  booktitle = {Advances in Neural Information Processing Systems},
  volume    = {35},
  pages     = {10078--10093},
  year      = {2022},
  doi       = {10.52202/068431-0732}
}

@inproceedings{bachmann2022multimae,
  author    = {Bachmann, Roman and Mizrahi, David and Atanov, Andrei and Zamir, Amir},
  title     = {MultiMAE: Multi-modal Multi-task Masked Autoencoders},
  booktitle = {Computer Vision -- ECCV 2022},
  series    = {Lecture Notes in Computer Science},
  volume    = {13697},
  pages     = {348--367},
  publisher = {Springer},
  year      = {2022},
  doi       = {10.1007/978-3-031-19836-6_20}
}

@inproceedings{nguyen2023climax,
  title     = {{C}lima{X}: A foundation model for weather and climate},
  author    = {Nguyen, Tung and Brandstetter, Johannes and Kapoor, Ashish and Gupta, Jayesh K and Grover, Aditya},
  booktitle = {Proceedings of the 40th International Conference on Machine Learning},
  pages     = {25904--25938},
  year      = {2023},
  editor    = {Krause, Andreas and Brunskill, Emma and Cho, Kyunghyun and Engelhardt, Barbara and Sabato, Sivan and Scarlett, Jonathan},
  volume    = {202},
  series    = {Proceedings of Machine Learning Research},
  publisher = {PMLR},
  url       = {https://proceedings.mlr.press/v202/nguyen23a.html}
}

@inproceedings{ronneberger2015unet,
  author    = {Ronneberger, Olaf and Fischer, Philipp and Brox, Thomas},
  title     = {U-Net: Convolutional Networks for Biomedical Image Segmentation},
  booktitle = {Medical Image Computing and Computer-Assisted Intervention -- MICCAI 2015},
  series    = {Lecture Notes in Computer Science},
  volume    = {9351},
  pages     = {234--241},
  publisher = {Springer},
  year      = {2015},
  doi       = {10.1007/978-3-319-24574-4_28}
}

@inproceedings{vaswani2017attention,
  author    = {Vaswani, Ashish and Shazeer, Noam and Parmar, Niki and Uszkoreit, Jakob and Jones, Llion and Gomez, Aidan N. and Kaiser, {\L}ukasz and Polosukhin, Illia},
  title     = {Attention Is All You Need},
  booktitle = {Advances in Neural Information Processing Systems},
  volume    = {30},
  pages     = {5998--6008},
  year      = {2017}
}

@article{su2024roformer,
  author  = {Su, Jianlin and Lu, Yu and Pan, Shengfeng and Murtadha, Ahmed and Wen, Bo and Liu, Yunfeng},
  title   = {RoFormer: Enhanced transformer with Rotary Position Embedding},
  journal = {Neurocomputing},
  volume  = {568},
  pages   = {127063},
  year    = {2024},
  doi     = {10.1016/j.neucom.2023.127063}
}

@article{takizuka1977,
  author  = {Takizuka, Tomonor and Abe, Hirotada},
  title   = {A binary collision model for plasma simulation with a particle code},
  journal = {Journal of Computational Physics},
  volume  = {25},
  number  = {3},
  pages   = {205--219},
  year    = {1977},
  doi     = {10.1016/0021-9991(77)90099-7}
}

@article{dorfman2008,
  author  = {Dorfman, S. and Daughton, W. and Roytershteyn, V. and Ji, H. and Ren, Y. and Yamada, M.},
  title   = {Two-dimensional fully kinetic simulations of driven magnetic reconnection with boundary conditions relevant to the Magnetic Reconnection Experiment},
  journal = {Physics of Plasmas},
  volume  = {15},
  number  = {10},
  pages   = {102107},
  year    = {2008},
  doi     = {10.1063/1.2991361}
}

\appendix

\section{VPIC corpus and training details}
\label{app:vpic}

The $\beta_{e0}=0.2$ corpus used in this work contains the Cartesian product of the run-name parameters shown in Table~\ref{tab:vpic_params}, yielding 250 complete simulations. This is one slice of the broader factorial VPIC scan; the present machine-learning experiments use only this fixed-$\beta_{e0}$ slice. The HDF5 packer stores the native VPIC outputs using processed channel names $(B_x,B_y,B_z,\rho)\equiv(B_R,B_\phi,B_Z,n_e/n_0)$ and records the run parameters as HDF5 attributes. The 90/10 run-level split uses seed 1234, producing 225 training runs and 25 disjoint validation runs.

\begin{table}[h]
    \centering
    \small
    \caption{Physical meaning and coverage of the VPIC parameters used for training and evaluation. The metadata field \texttt{dt} is an output cadence, not the PIC integration timestep.}
    \label{tab:vpic_params}
    \begin{tabular}{p{0.13\linewidth}p{0.43\linewidth}p{0.16\linewidth}p{0.19\linewidth}}
        \toprule
        Quantity & Definition & Units / normalization & Values \\
        \midrule
        $\beta_{e0}$ & Reference electron beta, $2\mu_0 n_0T_{e0}/B_0^2$ & dimensionless & 0.2 \\
        $\nu$ & Multiplier on the Coulomb collision rate evaluated at initial $n_0,T_{e0}$ & dimensionless & 0, 1, 2, 3, 5, 6, 7, 8, 9, 10 \\
        $B_{z0}$ & Uniform axial field added to the coil vacuum field; reconnecting-plane component, not a guide field & $B_0$ & 0, 0.1, 0.15, 0.2, 0.3 \\
        $\tau$ & Flux-core-current drive time & $\omega_{ci0}^{-1}$ & 40, 70, 100, 150, 200 \\
        \texttt{dt} & Output cadence $\Delta_{\rm out}$ & $\omega_{ci0}^{-1}$ & $\simeq 2$ \\
        Grid & Processed spatial grid & -- & $154\times62$ \\
        Domain & Retained reconstruction window & cm & $x\in[-50,50]$, $z\in[-21,21]$ \\
        \bottomrule
    \end{tabular}
\end{table}

\subsection{Simulation setup and normalization}

The VPIC calculations are two-dimensional and axisymmetric in cylindrical $(R,\phi,Z)$ geometry, with one cell in the toroidal direction and all three field and velocity components retained. The vessel spans $5\le R\le155$~cm and 150~cm in $Z$. Perfectly conducting walls reflect particles. Two flux cores of radius 16~cm are centered at $R=80$~cm and $Z=\pm53.5$~cm; each carries a prescribed toroidal current in its inner 4~cm and absorbs particles that reach the core. At $t=0$, the region outside the cores contains uniform Maxwellian electrons and ions with $m_i/m_e=25$, $T_{i0}=T_{e0}$, and no drift. The plasma is initialized in the coil vacuum field plus the uniform axial field $B_{z0}$, with $B_\phi=0$ initially.

Reconnection is driven by the prescribed flux-core current
\begin{equation}
    I(t)=I_0\,\frac{1+5\cos^2(\pi t/2\tau)}{6},
\end{equation}
which decreases from $I_0$ to $I_0/6$ over $0\le t\le\tau$ and then rises again; each nominal run extends to $t=1.5\tau$. This follows the MRX-type pull-reconnection setup of \citet{dorfman2008}. Coulomb electron--electron, electron--ion, and ion--ion collisions are included using the Takizuka--Abe binary-collision method \citep{takizuka1977}. The simulations use $\omega_{pe}/\omega_{ce0}=1$, and lengths are anchored to $n_0=2\times10^{13}\ {\rm cm^{-3}}$ for hydrogen, giving $d_i=5.09$~cm and $d_e=1.02$~cm. The native grid spacing is two electron Debye lengths, with 200 macro-particles per cell per species.

Frames are written at an output cadence $\Delta_{\rm out}\simeq2\,\omega_{ci0}^{-1}$. Each saved frame is a short boxcar time average over the first 20\% of its output interval. The PIC integration timestep is distinct from this cadence and is set independently at 0.98 of the Courant limit. The native output window used here is $30\le R\le130$~cm and $|Z|\le21$~cm; the processed radial coordinate is $x\equiv R-80~\mathrm{cm}$. The outputs contain $B_R$, $B_\phi$, and $B_Z$ in units of $B_0$ and electron density $n_e$ in units of $n_0$.

\subsection{Preprocessing, windowing, and split}

The preprocessing pipeline starts from four VPIC CSV sequences corresponding to native $B_R$, $B_\phi$, $B_Z$, and $n_e/n_0$, represented in the processed arrays as $B_x$, $B_y$, $B_z$, and $\rho$. For each simulation run, frames are matched by integer frame identifier; packing requires the four variables to contain identical frame identifiers, the first frame to be zero, and the sequence to be contiguous. Each frame is read as \,\texttt{float32} and stored in HDF5 with dataset layout $(T,4,X,Z)$ and channel order $(B_x,B_y,B_z,\rho)$. Run metadata, including the run-name field $\beta$ (physically $\beta_{e0}$), $\nu$, $B_{z0}$, $dt$, $\tau$, $T$, $X$, and $Z$, are stored as HDF5 attributes.

The training dataset lazily reads the HDF5 files and extracts every valid contiguous 24-frame window with start-time stride two, returning tensors in $(C,T,X,Z)$ order. Masking is not performed in the dataset loader; it is sampled online by the training loop after normalization. The split is made by complete run name with seed 1234, using 10\% of the 250 $\beta_{e0}=0.2$ runs for validation. This yields 225 training runs (7,059 windows) and 25 disjoint validation runs (591 windows). Channel means and standard deviations are estimated from training data only, with the training script configured to use at most 50 training batches for this estimate; the resulting statistics are the values reported in Section~3.1.

\subsection{Mask sampling and standardized evaluation}

During training, the magnetic field and density modalities independently draw one of the five mask families with equal probability. For \texttt{spatial\_random} and \texttt{spatial\_grid}, the exact visible-site count is sampled independently for each modality using a three-component mixture: probability 0.05 at zero visible sites, probability 0.05 at all $XZ$ sites, and otherwise an approximately log-uniform count in $[1,XZ-1]$. The three magnetic components share one magnetic layout. Spatial block masks draw a target block area from a continuous uniform mask fraction, randomize block location and aspect ratio, and choose inpainting versus outpainting with equal probability. Temporal random and temporal block masks draw the number of visible frames uniformly from $\{0,\ldots,T\}$; the former scatters the visible frames and the latter uses a contiguous visible or hidden interval. Magnetic and density pattern, severity, and layout are sampled independently.

The standardized benchmark in Table~\ref{tab:standardized} uses deterministic approximately 50\%-visible masks shared by all four channels, so differences primarily reflect geometry rather than severity. \texttt{spatial\_random} uses exactly $\lfloor XZ/2\rfloor$ distinct visible sites sampled with seed 4321; \texttt{spatial\_grid} uses the checkerboard rule $(x+z)\bmod 2=0$; \texttt{spatial\_block} keeps the lower half of the $x$ index visible and masks the upper half; \texttt{temporal\_random} keeps even-numbered local frames and masks odd-numbered frames; and \texttt{temporal\_block} keeps the first half of the 24-frame window and masks the second half.

Each evaluation window is scored separately. For each physical simulation run and mask setting, window-level metrics are first averaged arithmetically over all windows from that run. The reported central value is then the median of these per-run means across the 25 held-out runs, and the uncertainty band or interval is the 16th--84th percentile across runs. Overall MSE is the mean squared residual over all four standardized channels and all $(T,X,Z)$ locations. Density NRMSE is $\sqrt{\mathrm{mean}[(\hat{\rho}_{\rm std}-\rho_{\rm std})^2]}$. The $J_y$ metric is computed after inverse transformation to physical magnetic field units and is normalized by the training-set $J_y$ standard deviation as described in Section~5.

\subsection{Optimization and execution}

The final run uses AdamW with learning rate $2\times10^{-4}$, weight decay $10^{-4}$, 10 epochs of linear warmup, cosine decay to $2\times10^{-6}$, gradient-norm clipping at 1.0, and automatic mixed precision. Training runs for 4,500 epochs with batch size 8 per GPU on 16 GPUs, giving global batch size 128. Distributed execution uses PyTorch DistributedDataParallel with NCCL on four Perlmutter nodes with four NVIDIA A100-SXM4-80GB GPUs per node. After every epoch, the same held-out validation loader is evaluated under all five standardized mask patterns; no validation-skipping interval is used, so training contains 4,500 complete in-training validation passes. The run starts from random model weights; no previous checkpoint is used for initialization. Checkpoints record the model state, optimizer state, epoch, best validation score, training arguments, channel statistics, model version, and masking version. The launcher permits exact resume from the most recent checkpoint only when the recorded training parameters match the requested configuration.

For clarity, the effective model and training options are:
\begin{verbatim}
--betas 0.2 --delta-t 24 --stride-t 2
--batch-size 8 --epochs 4500 --lr 2e-4
--warmup-epochs 10 --min-lr 2e-6 --weight-decay 1e-4
--base-channels 24 --channel-mults 1 2 4 8
--use-attention --spatial-only-pooling --amp
--mask-patterns spatial_random 1 spatial_grid 1 spatial_block 1
                temporal_random 1 temporal_block 1
--density-visible-p-zero 0.05 --density-visible-p-full 0.05
--magnetic-visible-p-zero 0.05 --magnetic-visible-p-full 0.05
\end{verbatim}

The final run used the NERSC \texttt{pytorch/2.13.0} software environment. Runtime inspection of that installation gives Python 3.12.13, PyTorch \texttt{2.13.0+cu130}, CUDA 13.0 as reported by \texttt{torch.version.cuda}, NCCL 2.29.7, and cuDNN 9.20.0. The launcher invokes \texttt{python -m torch.distributed.run} (torchrun) with the \texttt{c10d} rendezvous backend and initializes PyTorch DistributedDataParallel with NCCL. Automatic mixed precision is enabled. These version strings are properties of the loaded NERSC module rather than hard-coded source constants.

\begin{table}[h]
    \centering
    \small
    \caption{Software and distributed-training environment used for the final run.}
    \label{tab:software_env}
    \begin{tabular}{ll}
        \toprule
        Item & Version / setting \\
        \midrule
        System & NERSC Perlmutter, A100-SXM4-80GB GPU nodes \\
        NERSC module & \texttt{pytorch/2.13.0} \\
        Python & 3.12.13 \\
        PyTorch & \texttt{2.13.0+cu130} \\
        CUDA build & 13.0 \\
        NCCL & 2.29.7 \\
        cuDNN & 9.20.0 \\
        Distributed training & PyTorch DDP / NCCL, 16 GPUs \\
        Precision & Automatic mixed precision (AMP) \\
        \bottomrule
    \end{tabular}
\end{table}

Training was continued across four scheduler sessions from the most recent checkpoint; resume restores the optimizer and epoch state and is accepted only when the recorded training configuration matches the requested run. Active distributed steps sum to 45,143~s (12~h~32~min), or approximately 200.6 allocated GPU-hours on 16 GPUs; including scheduler overhead outside the active steps, the four sessions span 47,072~s (13~h~4~min). These timing records include the validation pass performed after every epoch; no separate train-only and validation-only timers were recorded. The approximately 200 GPU-hour figure refers to the final reported training run; preliminary model-development runs used additional compute that was not comprehensively tracked, and separate wall-clock timings were not recorded for the post-hoc evaluation scripts.

\subsection{Reproducibility scope}

The released training and evaluation code specifies the machine-learning pipeline starting from the processed VPIC HDF5 field outputs, including window construction, masking, training, checkpointing, and evaluation. The original VPIC simulations were generated independently of the present machine-learning study, and the complete VPIC input decks and numerical configuration are not included in the current release. An anonymized code and checkpoint archive is available at \url{https://anonymous.4open.science/r/neurips_anonymous_supplement-11CC/}. The archive provides the training and evaluation code, model implementation, exact run-level split metadata, sanitized final checkpoint, configuration files, software-environment specification, and reproduction scripts. The processed VPIC corpus itself is not distributed with the present submission while data provenance, attribution, and release permissions are being finalized with the simulation contributors. Consequently, the released materials allow the model implementation, training configuration, checkpoint, and evaluation pipeline to be inspected and reused, but exact re-execution of the reported VPIC numerical evaluations requires access to the processed corpus. Regenerating the underlying VPIC simulations from particle-in-cell initial conditions is not required once the processed corpus is available.

\section{Architecture and implementation details}
\label{app:architecture}

Table~\ref{tab:architecture_shapes} summarizes the tensor geometry of the model for one $24\times154\times62$ window, while Table~\ref{tab:param_breakdown} reports the exact trainable parameter allocation. MaxPool3d uses kernel=stride $(1,2,2)$, so odd spatial dimensions are floored on the encoder path. The decoder uses trilinear interpolation to the exact skip-tensor shape before concatenation, restoring the original $154\times62$ grid without requiring divisibility by powers of two. The two attention branches together contain 186,048 parameters (3.37\% of the model); most parameters remain in the convolutional encoder, bottleneck, and decoder.

\begin{table}[!ht]
    \centering

    \begin{minipage}[t]{0.59\linewidth}
        \centering
        \scriptsize
        \caption{Stage-by-stage feature geometry of the residual U-Net. Shapes omit batch dimension and are written as channels $\times T\times X\times Z$.}
        \label{tab:architecture_shapes}
        \resizebox{\linewidth}{!}{%
        \begin{tabular}{lll}
            \toprule
            Stage & Feature shape & Attention tokens \\
            \midrule
            Input & $8\times24\times154\times62$ & -- \\
            Encoder 0 & $24\times24\times154\times62$ & -- \\
            Encoder 1 & $48\times24\times77\times31$ & -- \\
            Encoder 2 + attention & $96\times24\times38\times15$ & 13,680 \\
            Encoder 3 & $192\times24\times19\times7$ & -- \\
            Bottleneck + attention & $192\times24\times19\times7$ & 3,192 \\
            Decoder 2 & $96\times24\times38\times15$ & -- \\
            Decoder 1 & $48\times24\times77\times31$ & -- \\
            Decoder 0 & $24\times24\times154\times62$ & -- \\
            Output & $4\times24\times154\times62$ & -- \\
            \bottomrule
        \end{tabular}%
        }
    \end{minipage}
    \hfill
    \begin{minipage}[t]{0.38\linewidth}
        \centering
        \scriptsize
        \caption{Trainable parameter count by top-level module.}
        \label{tab:param_breakdown}
        \resizebox{\linewidth}{!}{%
        \begin{tabular}{lr}
            \toprule
            Module & Parameters \\
            \midrule
            Encoder 0 & 21,000 \\
            Encoder 1 & 94,608 \\
            Encoder 2 & 378,144 \\
            Encoder 3 & 1,512,000 \\
            Bottleneck residual block & 1,991,040 \\
            Decoder 2 & 1,023,264 \\
            Decoder 1 & 255,888 \\
            Decoder 0 & 64,008 \\
            Encoder-2 attention & 37,440 \\
            Bottleneck attention & 148,608 \\
            Output $1\times1\times1$ conv. & 100 \\
            \midrule
            Total & 5,526,100 \\
            \bottomrule
        \end{tabular}%
        }
    \end{minipage}
\end{table}

The two attention modules use the same structural sequence,
\[
    \mathbf{u}
    \mapsto
    \mathbf{u}
    +
    W_o\,\mathrm{SDPA}
    \!\left(
      \mathrm{RoPE}(Q),\mathrm{RoPE}(K),V
    \right),
\]
with $Q,K,V$ obtained from a single linear projection of $\mathrm{LayerNorm}(\mathbf{u})$. At Encoder 2, $C=96$, $H=8$, and $d_h=12$, so each of $Q,K,V$ has 96 features per token and the combined projection has 288 outputs. At the bottleneck, $C=192$, $H=8$, and $d_h=24$, giving 192 features per $Q$, $K$, or $V$ tensor and 576 outputs in the combined projection. The RoPE slices are $(4,4,4)$ dimensions per head for $(t,x,z)$ at Encoder 2 and $(8,8,8)$ at the bottleneck. No MLP/FFN or attention dropout is used.

The implementation uses \texttt{torch.nn.functional.scaled\_dot\_product\_attention} with \texttt{dropout\_p=0} and \texttt{is\_causal=False}. No attention backend is explicitly selected in the model code. The residual attention output projection and the final field-output convolution are both initialized to zero, so the added attention path and the final learned field correction start from zero.

\section{Long-sequence sliding reconstruction}
\label{app:sliding}

The network is trained on 24-frame windows. To test longer sequences without retraining, we recursively apply the model with sliding steps of 1, 12, and 24 frames. The appendix experiment uses 1,000 fixed density probes (10.47\% spatial visibility), considers the first 48 frames of each sufficiently long held-out run, and evaluates 18 validation runs. The statistical unit is the simulation run: each run contributes one 48-frame RMSE curve per setting, after which median and 16th--84th percentile bands are computed across runs.

\begin{figure}[h]
    \centering
    \includegraphics[width=\linewidth]{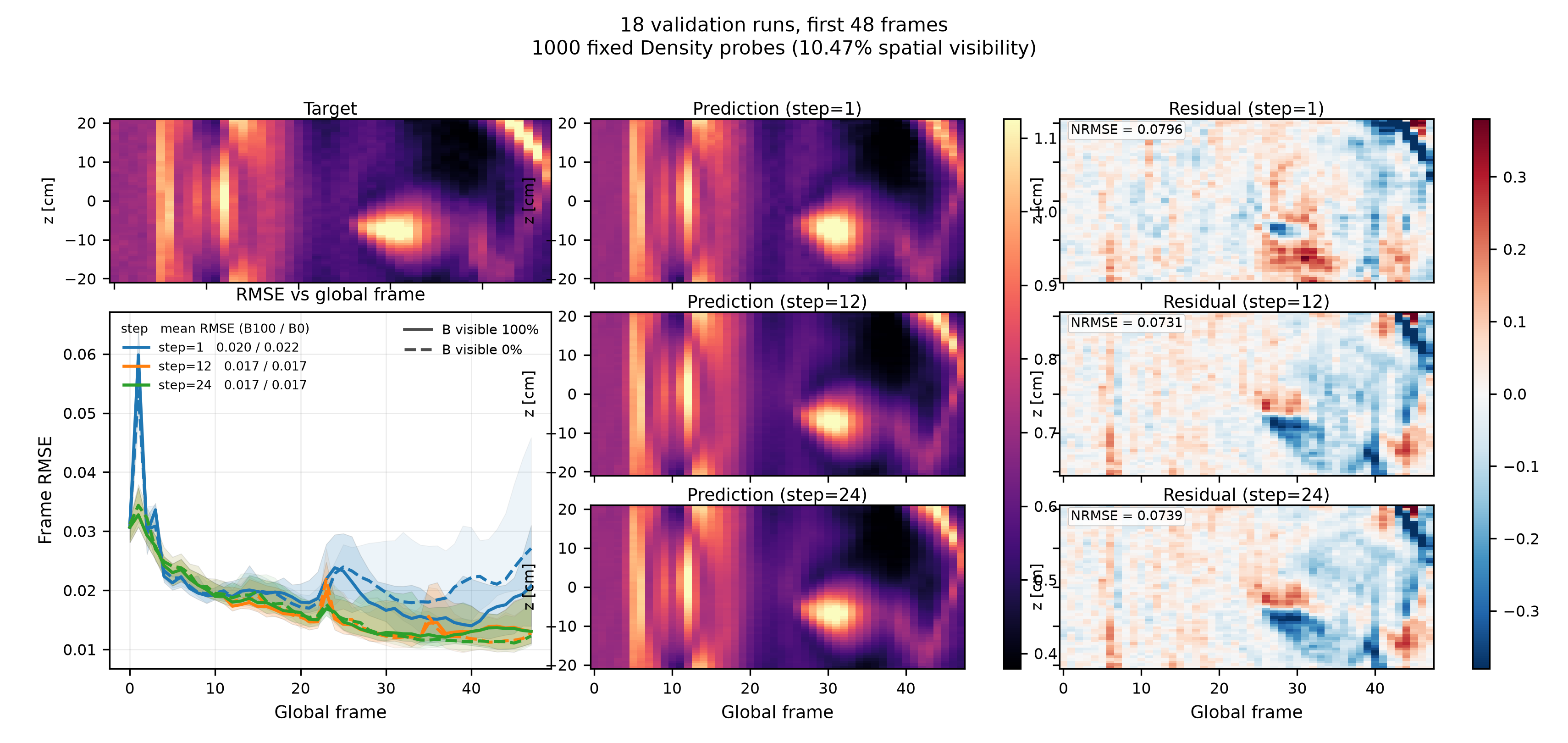}
    \caption{Sliding-window extension to 48-frame sequences with 1,000 fixed density probes. Left: median frame RMSE across 18 held-out runs for different slide steps and magnetic conditions. Middle/right: predictions and residuals for a representative B-hidden trajectory.}
    \label{fig:sliding}
\end{figure}

Figure~\ref{fig:sliding} shows that more overlap is not necessarily better. The time-averaged median frame RMSE is approximately $0.020/0.022$ for step 1, compared with $0.017/0.017$ for steps 12 and 24 (fully visible / fully hidden magnetic field). On the displayed B-hidden canonical trajectory, aggregate 48-frame NRMSE values are 0.0796, 0.0731, and 0.0739 for steps 1, 12, and 24, respectively. This suggests that extremely frequent recursive reuse can accumulate model error rather than monotonically refine the trajectory. The procedure is an offline long-sequence reconstruction method, not a causal forecasting algorithm.

\newpage

\end{document}